# A Structured Prediction Approach for Label Ranking

**Anna Korba**  anna.korba@telecom-paristech.fr
**Alexandre Garcia**  garcia@telecom-paristech.fr
**Florence d'Alché-Buc**  florence.dalche@telecom-paristech.fr
*LTCI, Télécom ParisTech, Université Paris-Saclay*
*Paris, France*

## Abstract

We propose to solve a label ranking problem as a structured output regression task. We adopt a least square surrogate loss approach that solves a supervised learning problem in two steps: the regression step in a well-chosen feature space and the pre-image step. We use specific feature maps/embeddings for ranking data, which convert any ranking/permutation into a vector representation. These embeddings are all well-tailored for our approach, either by resulting in consistent estimators, or by solving trivially the pre-image problem which is often the bottleneck in structured prediction. We also propose their natural extension to the case of partial rankings and prove their efficiency on real-world datasets.

## 1. Introduction

Label ranking is a prediction task which aims at mapping input instances to a (total) order over a given set of labels indexed by $\{1,\ldots,K\}$. This problem is motivated by applications where the output reflects some preferences, or order or relevance, among a set of objects: in pattern recognition (see Geng and Luo (2014)), the goal is to predict the different objects which are the more likely to appear in an image among a predefined set; similarly, in sentiment analysis, (see Wang et al. (2011)) the prediction of the emotions expressed in a document is cast as a label ranking problem over a set of possible affective expressions. Another application is metalearning, where the goal is to rank a set of algorithms according to their suitability based on the characteristics of a target dataset and learning problem (see Aiguzhinov et al. (2010), Brazdil et al. (2003)).

More formally, the goal of label ranking is to map a vector $x$ lying in some feature space $\mathcal{X}$ to a ranking $y$ lying in the space of rankings $\mathcal{Y}$. A ranking is an order on $\{1,\ldots,K\}$, i.e a binary relation verifying transitivity. These relations linking the components of the $y$ objects induce a structure on the output space $\mathcal{Y}$. The label ranking task thus naturally enters the framework of structured output prediction for which an abundant litterature is available Nowozin and Lampert (2011). In this paper, we adopt the *surrogate least square loss* approach introduced in the context of output kernels Cortes et al. (2005); Kadri et al. (2013); Brouard et al. (2016) and recently theoretically studied by Ciliberto et al. (2016); Osokin et al. (2017) using Calibration theory Steinwart and Christmann (2008). This approach divides the learning task in two steps: the first one is a vector regression step in a Hilbert space where the outputs objects are represented, and the second one solves a pre-image problem to retrieve an output object in the $\mathcal{Y}$ space. In this framework, the algorithmic performances of the learning and prediction tasks and the generalization properties of the resulting predictor crucially rely on some properties of the output objects representation. In this work we highlight the properties of some embeddings dedicated to ranking data.

Our contribution are three folds: (1) we cast the label ranking problem into the structured prediction framework and propose embeddings dedicated to ranking representation, (2) for each embedding we propose a solution to the pre-image problem and study its algorithmic complexity and (3) we provide theoretical and empirical evidence for the relevance of our method.

The paper is organized as follows. In section 2, the main definitions and notations of objects considered through the paper are introduced, and section 3 is devoted to the setting of the learning problem and statistical framework considered. Section 4 describes at length the embeddings we propose and section 5 details the theoretical and computational advantages of our approach. Finally section 6 contains empirical results on benchmark datasets.

## 2. Preliminaries

### 2.1 Mathematical background and notations

Consider a set of items indexed by $\{1, \ldots, K\}$, that we will denote $[\![K]\!]$. Rankings can be complete (i.e, involving all the items) or incomplete and for both cases, they can be without-ties (total order) or with-ties (weak order). A full ranking is a complete, and without-ties ranking of the items in $[\![K]\!]$. It can be seen as a permutation, i.e a bijection $\sigma : [\![K]\!] \to [\![K]\!]$, mapping each item $i$ to its rank $\sigma(i)$. The rank of item $i$ is thus $\sigma(i)$ and the item ranked at position $j$ is $\sigma^{-1}(j)$. We say that $i$ is preferred over $j$ (denoted by $i \succ j$) according to $\sigma$ if and only if $i$ is ranked lower than $j$: $\sigma(i) < \sigma(j)$. The set of all permutations over $K$ items is the symmetric group which we denote by $\mathfrak{S}_K$. A partial ranking is a complete ranking including ties, and is also referred as a weak order or bucket order in the litterature (see Kenkre et al. (2011)). This includes in particular the top-$k$ rankings, that is to say partial rankings dividing items in two groups, the first one being the $k \leq K$ most relevant items and the second one including all the rest. These top-k rankings are given a lot of attention because of their relevance for modern applications, especially search engines or recommendation systems (see Ailon (2010)). An incomplete ranking is a strict order involving only a small subset of items, and includes as a particular case pairwise comparisons, also very relevant in large-scale settings for ranking when the number of items is very large. We now introduce the main notations used through the paper. For any function $f$, $Im(f)$ denotes the image of $f$, and $f^{-1}$ its inverse. The indicator function of any event $\mathcal{E}$ is denoted by $\mathbb{I}\{\mathcal{E}\}$. We will denote by $sign$ the function such that for any $x \in \mathbb{R}$, $sign(x) = \mathbb{I}\{x > 0\} - \mathbb{I}\{x < 0\}$. The notations $\|.\|$ and $|.|$ denote respectively the usual $l_2$ and $l_1$ norm in a euclidean space. Finally, for any integers $a \leq b$, $[\![a, b]\!]$ denotes the set $\{a, a+1, \ldots, b\}$, and for any finite set $C$, $\#C$ denotes its cardinality.

### 2.2 Related work

An overview of label ranking algorithms can be found in Vembu and Gärtner (2010), Zhou et al. (2014)), but we recall here the main contributions. One of the first approaches, called *pairwise classification* (see Fürnkranz and Hüllermeier (2003)) has been to transform the label ranking problem into $K(K-1)/2$ binary classification problems. For each possible pair of labels $1 \leq i < j \leq K$, the authors learn a model $m_{ij}$ that decides for any given example whether $i \succ j$ or $j \succ i$ holds. The model is trained with all examples for which either $i \succ j$ or $j \succ i$ is known (all examples for which nothing is known about this pair are ignored). At classification time, an example is submitted to all $K(K-1)/2$ theories, and each prediction is interpreted as a vote for a label: if the classifier $m_{ij}$ predicts $i \succ j$, this counts as a vote for label $i$. The labels are ranked according to the number of votes. Then, a large part of the dedicated literature was devoted to adapting classical partitioning methods such as k-nearest neighbors (see Zhang and Zhou (2007), Chiang et al. (2012)) or tree-based methods, in a parametric (Cheng et al. (2010), Cheng et al. (2009), Aledo et al. (2017)) or a non-parametric way (see Cheng and Hüllermeier (2013), Yu et al. (2010), Zhou and Qiu (2016), Clémençon et al. (2017), Sá et al. (2017)). Finally, some approaches are rule-based (see Gurrieri et al. (2012), de Sá et al. (2018)). We will compare our numerical results with the best performances attained by these methods on a set of benchmark datasets of the label ranking problem in section 6.

## 3. Structured prediction for label ranking

### 3.1 Learning problem

Our goal is to learn a function $s : \mathcal{X} \to \mathcal{Y}$ between a feature space $\mathcal{X}$ and a structured output space $\mathcal{Y}$, that we set to be $\mathfrak{S}_K$ the space of full rankings over the set of items $[\![K]\!]^1$. The quality of a prediction $s(x)$ is measured using a loss function $\Delta : \mathfrak{S}_K \times \mathfrak{S}_K \to \mathbb{R}$, where $\Delta(s(x), \sigma)$ is the cost suffered by predicting $s(x)$ for the true output $\sigma$. We suppose that the input/output pairs $(x, \sigma)$ come from some fixed distribution $P$ on $\mathcal{X} \times \mathfrak{S}_K$. The label ranking problem is then defined as:

$$\text{minimize}_{s:\mathcal{X} \to \mathfrak{S}_K} \mathcal{E}(s), \quad \text{with} \quad \mathcal{E}(s) = \int_{\mathcal{X} \times \mathfrak{S}_K} \Delta(s(x), \sigma) dP(x, \sigma). \tag{1}$$

In this paper, we propose to study how to solve this problem and its empirical counterpart for a family of loss functions based on some ranking embedding $\phi : \mathfrak{S}_K \to \mathcal{F}$ that maps the permutations $\sigma \in \mathfrak{S}_K$ into a Hilbert space $\mathcal{F}$:

$$\Delta(\sigma, \sigma') = \|\phi(\sigma) - \phi(\sigma)\|_{\mathcal{F}}^2. \tag{2}$$

This loss presents two main advantages: first, there exists popular losses for ranking data that can take this form within a finite dimensional Hilbert space $\mathcal{F}$, second, this choice benefits from the theoretical results on Surrogate Least Square problems for Structured Prediction using Calibration theory Ciliberto et al. (2016) and of works of Brouard et al. (2016) on Structured Output prediction within vector-valued Reproducing Kernel Hilbert Spaces. These works approach Structured Output Prediction along a common angle by introducing a surrogate problem involving a function $g : \mathcal{X} \to \mathcal{F}$ (with values in $\mathcal{F}$) and a surrogate loss $L(g(x), \sigma)$ to be minimized instead of Eq. 1. In the context of true risk minimization, the surrogate problem for our case writes as:

$$\text{minimize }_{g:\mathcal{X} \to \mathcal{F}} \mathcal{R}(g), \quad \text{with} \quad \mathcal{R}(g) = \int_{\mathcal{X} \times \mathfrak{S}_K} L(g(x), \phi(\sigma)) dP(x, \sigma). \tag{3}$$

---
1. In Section 4.4, we will extend the task to partial and incomplete rankings

with the following surrogate loss:

$$L(g(x), \phi(\sigma)) = \|g(x) - \phi(\sigma)\|_{\mathcal{F}}^2. \tag{4}$$

Problem of Eq. (3) is in general easier to optimize since $g$ has values in $\mathcal{F}$ instead of the set of structured objects $\mathcal{Y}$, here $\mathfrak{S}_K$. The solution of (3), denoted as $g^*$, can be written for any $x \in \mathcal{X}$: $g^*(x) = \mathbb{E}[\phi(\sigma)|x]$. Eventually, a candidate $s(x)$ pre-image for $g^*(x)$ can then be obtained by solving:

$$s(x) = \operatorname*{argmin}_{\sigma \in \mathfrak{S}_K} L(g^*(x), \phi(\sigma)) \tag{5}$$

In the context of Empirical Risk Minimization, a training sample $\{(x_i, \sigma_i), i = 1, \ldots N\}$, with $N$ i.i.d. copies of the random variable $(x, \sigma)$ is available. The Surrogate Least Square approach for Label Ranking Prediction decomposes into two steps:

- Step 1: minimize a regularized empirical risk to provide an estimator of the minimizer of the regression problem in Eq. (3):

$$\operatorname*{minimize}_{g \in \mathcal{H}} \mathcal{R}_\mathcal{S}(g), \text{ with } \mathcal{R}_\mathcal{S}(g) = \frac{1}{N} \sum_{i=1}^N L(g(x_i), \phi(\sigma_i)) + \Omega(g). \tag{6}$$

  with an appropriate choice of hypothesis space $\mathcal{H}$ and complexity term $\Omega(g)$. We denote by $\widehat{g}$ a solution of (6).

- Step 2: solve, for any $x$ in $\mathcal{X}$, the pre-image problem that provides a prediction in the original space $\mathfrak{S}_K$:

$$\widehat{s}(x) = \operatorname*{argmin}_{\sigma \in \mathfrak{S}_K} \|\phi(\sigma) - \widehat{g}(x)\|_{\mathcal{F}}^2 \tag{7}$$

  The pre-image operation can be written as $\widehat{s}(x) = d \circ \widehat{g}(x)$ with $d$ the decoding function:

$$d(h) = \operatorname*{argmin}_{\sigma \in \mathfrak{S}_K} \|\phi(\sigma) - h\|_{\mathcal{F}}^2 \text{ for all } h \in \mathcal{F} \tag{8}$$

  applied on $\widehat{g}$ for any $x \in \mathcal{X}$.

This paper studies how to leverage the choice of the embedding $\phi$ to obtain a good compromise between computational complexity and theoretical guarantees. Typically, the pre-image problem on the discrete set $\mathfrak{S}_K$ can be eased for appropriate choices of $\phi$ as we show in section 4, leading to efficient solutions. In the same time, one would like to benefit from theoretical guarantees and control the excess risk of the proposed predictor.

In the following subsection we exhibit popular losses for ranking data that we will use for the label ranking problem.

### 3.2 Losses for ranking

We now present losses $\Delta$ on $\mathfrak{S}_K$ that we will consider for the label ranking task. A natural loss for full rankings, i.e. permutations in $\mathfrak{S}_K$, is a distance between permutations. Several distances on $\mathfrak{S}_K$ are widely used in the literature, one of the most popular being the Kendall's $\tau$ distance, which counts the number of pairwise disagreements between two permutations $\sigma, \sigma' \in \mathfrak{S}_K$:

$$\Delta_\tau(\sigma, \sigma') = \sum_{i<j} \mathbb{I}[(\sigma(i) - \sigma(j))(\sigma'(i) - \sigma'(j)) < 0]. \tag{9}$$

The maximal Kendall's $\tau$ distance is thus $K(K-1)/2$, the total number of pairs. Another well-spread distance between permutations is the Hamming distance, which counts the number of entries on which two permutations $\sigma, \sigma' \in \mathfrak{S}_K$ disagree:

$$\Delta_H(\sigma, \sigma') = \sum_{i=1}^{K} \mathbb{I}[\sigma(i) \neq \sigma'(i)]. \tag{10}$$

The maximal Hamming distance is thus $K$, the number of labels or items. In the next section we show how these distances can be written as (2) for a well chosen embedding $\phi$.

## 4. Output embeddings for rankings

In what follows, we study three embeddings tailored to represent full rankings/permutations in $\mathfrak{S}_K$ and discuss their properties in terms of link with the ranking distances $\Delta_\tau$ and $\Delta_H$, and in terms of algorithmic complexity for the pre-image problem (5) induced.

### 4.1 The Kemeny embedding

Motivated by the minimization of the Kendall's $\tau$ distance $\Delta_\tau$, we study the Kemeny embedding, previously introduced for the ranking aggregation problem (see Jiao et al. (2016)):

$$\phi_\tau \colon \mathfrak{S}_K \to \mathbb{R}^{K(K-1)/2}$$
$$\sigma \mapsto (\text{sign}(\sigma(j) - \sigma(i)))_{1 \leq i < j \leq K} .$$

which maps any permutation $\sigma \in \mathfrak{S}_K$ into $Im(\phi_\tau) \subsetneq \{-1, 1\}^{K(K-1)/2}$ (that we have embedded into the Hilbert space $(\mathbb{R}^{K(K-1)/2}, \langle .,. \rangle)$). One can show that the square of the euclidean distance between the mappings of two permutations $\sigma, \sigma' \in \mathfrak{S}_K$ recovers their Kendall's $\tau$ distance (proving at the same time that $\phi_\tau$ is injective) up to a constant: $\|\phi_\tau(\sigma) - \phi_\tau(\sigma')\|^2 = 4\Delta_\tau(\sigma, \sigma')$. The Kemeny embedding then naturally appears to be a good candidate to build a surrogate loss related to $\Delta_\tau$. By noticing that $\phi_\tau$ has a constant norm ($\forall \sigma \in \mathfrak{S}_K, \|\phi_\tau(\sigma)\| = \sqrt{K(K-1)/2}$), we can rewrite the pre-image problem (7) under the form:

$$\widehat{s}(x) = \underset{\sigma \in \mathfrak{S}_K}{\operatorname{argmin}} -\langle \phi_\tau(\sigma), \widehat{g}(x) \rangle. \tag{11}$$

To compute (11), one can solve firstly an Integer Linear Program (ILP) to find $\widehat{\phi_\sigma} = \operatorname{argmin}_{\phi_\sigma \in Im(\phi_\tau)} -\langle \phi_\sigma, \widehat{g}(x) \rangle$, and then find the output object $\sigma = \phi_\tau^{-1}(\widehat{\phi_\sigma})$. The second step can be performed in $\mathcal{O}(K^2)$ by means of the Copeland method (see Merlin and Saari (1997)), which ranks the items by their number of pairwise victories[2]. In contrast, the ILP problem is harder to solve since it involves a minimization over $Im(\phi_\tau)$, a set of structured vectors since their coordinates are strongly correlated by the *transitivity* property of rankings. Indeed, consider a vector $v \in Im(\phi_\tau)$, so $\exists \sigma \in \mathfrak{S}_K$ such that $v = \phi_\tau(\sigma)$. Then, for any $1 \leq i < j < k \leq K$, if its coordinates corresponding to the pairs $(i,j)$ and $(j,k)$ are equal to one (meaning that $\sigma(i) < \sigma(j)$ and $\sigma(j) < \sigma(k)$), then the coordinate corresponding to the pair $(i,k)$ cannot contradict the others and must be set to one as well. Since $\phi_\sigma = (\phi_\sigma)_{i,j} \in Im(\phi_\tau)$ is only defined for $1 \leq i < j \leq K$, to encode the transitivity constraint we introduce $\phi'_\sigma = (\phi'_\sigma)_{i,j} \in \mathbb{R}^{K(K-1)}$ defined by $(\phi'_\sigma)_{i,j} = (\phi_\sigma)_{i,j}$ if $1 \leq i < j \leq K$ and $(\phi'_\sigma)_{i,j} = -(\phi_\sigma)_{i,j}$ else, and write the ILP problem as follows:

$$\widehat{\phi_\sigma} = \underset{\phi'_\sigma}{\operatorname{argmin}} \sum_{1 \leq i,j \leq K} \widehat{g}(x)_{i,j} (\phi'_\sigma)_{i,j},$$

$$s.c. \begin{cases} (\phi'_\sigma)_{i,j} \in \{-1, 1\} & \forall\, i, j \\ (\phi'_\sigma)_{i,j} + (\phi'_\sigma)_{j,i} = 0 & \forall\, i, j \\ -1 \leq (\phi'_\sigma)_{i,j} + (\phi'_\sigma)_{j,k} + (\phi'_\sigma)_{k,i} \leq 1 & \forall\, i, j, k \text{ s.t. } i \neq j \neq k. \end{cases} \quad (12)$$

---
2. Copeland method firstly affects a score $s_i$ for item $i$ as: $s_i = \sum_{j \neq i} \mathbb{I}\{\sigma(i) < \sigma(j)\}$ and then ranks the items by decreasing score.

Such a problem is NP-Hard but in practice, branch and bound algorithms find the solution in a reasonable time for a reduced number of labels $K$. We discuss the computational implications of choosing the Kemeny embedding section 5.2. We now turn to the study of an embedding devoted to build a surrogate loss for the Hamming distance.

### 4.2 The Hamming embedding

Another well-spread embedding for permutations, that we will call the Hamming embedding, consists in mapping $\sigma$ to its permutation matrix $\phi_H(\sigma)$:

$$\phi_H \colon \mathfrak{S}_K \to \mathbb{R}^{K \times K}$$
$$\sigma \mapsto (\mathbb{I}\{\sigma(i) = j\})_{1 \leq i,j \leq K} \,,$$

where we have embedded $Im(\phi_H) \subsetneq \{0,1\}^{K \times K}$[3] into the Hilbert space $(\mathbb{R}^{K \times K}, \langle .,. \rangle)$ with $\langle .,. \rangle$ the Froebenius inner product. This embedding shares similar properties with the Kemeny embedding: first, it is also of constant (Froebenius) norm, since $\forall \sigma \in \mathfrak{S}_K$, $\|\phi_H(\sigma)\| = \sqrt{K}$. Then, the square of the euclidean distance between the mappings of two permutations $\sigma, \sigma' \in \mathfrak{S}_K$ recovers their Hamming distance (proving that $\phi_H$ is also injective): $\|\phi_H(\sigma) - \phi_H(\sigma')\|^2 = \Delta_H(\sigma, \sigma')$. Once again, the *pre-image problem* consists in solving the linear program:

$$\widehat{s}(x) = \underset{\sigma \in \mathfrak{S}_K}{\operatorname{argmin}} -\langle \phi_H(\sigma), \widehat{g}(x) \rangle \quad (13)$$

which is, as for the Kemeny embedding previously, divided in a minimization step, i.e. find $\widehat{\phi_\sigma} = \operatorname{argmin}_{\phi_\sigma \in Im(\phi_H)} -\langle \phi_\sigma, \widehat{g}(x) \rangle$, and an inversion step, i.e. compute $\sigma = \phi_H^{-1}(\widehat{\phi_\sigma})$. The inversion step is of complexity $\mathcal{O}(K^2)$ since it involves scrolling through all the rows (items $i$) of the matrix $\widehat{\phi_\sigma}$ and all the columns (to find their positions $\sigma(i)$). The minimization step

---

3. The space $Im(\phi_H)$ is actually called the set of permutation matrices.

itself writes as the following problem:

$$\widehat{\phi_\sigma} = \underset{\phi_\sigma}{\mathrm{argmax}} \sum_{1 \leq i,j \leq K} \widehat{g}(x)_{i,j}(\phi_\sigma)_{i,j}$$
$$s.c \begin{cases} (\phi_\sigma)_{i,j} \in \{0,1\} & \forall\, i,j \\ \sum_i (\phi_\sigma)_{i,j} = \sum_j (\phi_\sigma)_{i,j} = 1 & \forall\, i,j \end{cases} \quad (14)$$

which can be solved with the Hungarian algorithm (see Kuhn (1955)) in $\mathcal{O}(K^3)$ time. Now we turn to the study of an embedding which presents efficient algorithmic properties.

### 4.3 Lehmer code

A permutation $\sigma = (\sigma(1), \ldots, \sigma(K)) \in \mathfrak{S}_K$ may be uniquely represented via its Lehmer code (also called the inversion vector), i.e. a word of the form $c_\sigma \in \mathcal{C}_K \triangleq \{0\} \times [\![0,1]\!] \times [\![0,2]\!] \times \cdots \times [\![0, K-1]\!]$, where for $j = 1, \ldots, K$:

$$c_\sigma(j) = \#\{i : i < j, \sigma(i) > \sigma(j)\} \quad (15)$$

The coordinate $c_\sigma(j)$ is thus the number of elements $i$ with index smaller than $j$ that are ranked higher than $j$ in the permutation $\sigma$. By default, $c_\sigma(1) = 0$ and is typically omitted. For instance, we have:

| e | 1 | 2 | 3 | 4 | 5 | 6 | 7 | 8 | 9 |
|---|---|---|---|---|---|---|---|---|---|
| $\sigma$ | 2 | 1 | 4 | 5 | 7 | 3 | 6 | 9 | 8 |
| $c_\sigma$ | 0 | 1 | 0 | 0 | 0 | 3 | 1 | 0 | 1 |

It is well known that the Lehmer code is bijective, and that the encoding and decoding algorithms have linear complexity $\mathcal{O}(K)$ (see Mareš and Straka (2007), Myrvold and Ruskey (2001)). This embedding has been recently used for ranking aggregation of full or partial rankings (see Li et al. (2017)). Our idea is thus to consider the following Lehmer mapping for label ranking;

$$\phi_L \colon \mathfrak{S}_K \to \mathbb{R}^K$$
$$\sigma \mapsto (c_\sigma(i)))_{i=1,\ldots,K} \,,$$

which maps any permutation $\sigma \in \mathfrak{S}_K$ into the space $\mathcal{C}_K$ (that we have embedded into the Hilbert space $(\mathbb{R}^K, \langle .,. \rangle)$). The loss function in the case of the Lehmer embedding is thus the following:

$$\Delta_L(\sigma, \sigma') = \|\phi_L(\sigma) - \phi_L(\sigma')\|^2 \quad (16)$$

Notice that $|\phi_L(\sigma)| = d_\tau(\sigma, e)$ where $e$ is the identity permutation, a quantity which is also called the number of inversions of $\sigma$. Therefore, in contrast to the previous mappings, the norm $\|\phi_L(\sigma)\|$ is not constant for any $\sigma \in \mathfrak{S}_K$. It is not possible to write the loss $\Delta_L(\sigma, \sigma')$ as $-\langle \phi_L(\sigma), \phi_L(\sigma') \rangle^4$ as for the previous embeddings. Moreover, this mapping is not distance preserving and it can be proven that $\frac{1}{K-1}\Delta_\tau(\sigma, \sigma') \leq |\phi_L(\sigma) - \phi_L(\sigma')| \leq \Delta_\tau(\sigma, \sigma')$ (see Wang et al. (2015)). However, the Lehmer embedding still enjoys great advantages. Firstly, its

---

4. The scalar product of two embeddings of two permutations $\phi_L(\sigma), \phi_L(\sigma')$ is not maximized for $\sigma = \sigma'$.

coordinates are decoupled, which will enable a trivial solving of the inverse image step (7). Indeed we can write explicitly its solution as:

$$\widehat{s}(x) = \underbrace{\phi_L^{-1} \circ d_L}_{d} \circ \widehat{g}(x) \quad \text{with} \quad \begin{array}{c} d_L \colon \mathbb{R}^K \to \mathcal{C}_K \\ (h_i)_{i=1,\dots,K} \mapsto (\underset{j \in [\![0, i-1]\!]}{\operatorname{argmin}} (h_i - j))_{i=1,\dots,K}. \end{array} \quad (17)$$

Then, there may be repetitions in the coordinates of the Lehmer embedding, allowing for a compact representation. Finally, as explained in the next subsection, it can be easily extended to partial rankings.

### 4.4 Extension to partial and incomplete rankings

In many real-world applications, one does not observe full rankings but only partial or incomplete rankings (see the definitions section 2.1). We now discuss to what extent the embeddings we propose for permutations can be adapted to this kind of rankings. Firstly, it appears that the Lehmer embedding can be generalized to partial ranking entries. Indeed, in Li et al. (2017), the authors propose a generalization of the Lehmer code for partial rankings. We recall that a tie in a ranking happens when $\#\{i \neq j, \sigma(i) = \sigma(j)\} > 0$. The generalized representation $c'$ takes into account ties, so that for any partial ranking $\widetilde{\sigma}$:

$$c'_{\widetilde{\sigma}}(j) = \#\{i : i < j, \widetilde{\sigma}(i) \geq \widetilde{\sigma}(j)\} \quad (18)$$

Clearly, $c'_{\widetilde{\sigma}}(j) \geq c_{\widetilde{\sigma}}(j)$ for all $j \in [\![K]\!]$. Given a partial ranking $\widetilde{\sigma}$, it is possible to break its ties to convert it in a permutation $\sigma$ as follows: for $i, j \in [\![K]\!]^2$, if $\widetilde{\sigma}(i) = \widetilde{\sigma}(j)$ then $\sigma(i) = \sigma(j)$ iff $i < j$. The entries $j = 1, \dots, K$ of the Lehmer codes of $\widetilde{\sigma}$ (see (19)) and $\sigma$ (see (15)) then verify:

$$c'_{\widetilde{\sigma}}(j) = c_\sigma(j) + IN_j - 1 \quad , \quad c_{\widetilde{\sigma}}(j) = c_\sigma(j) \quad (19)$$

where $IN_j = \#\{i \leq j, \widetilde{\sigma}(i) = \widetilde{\sigma}(j)\}$. An example illustrating the extension of the Lehmer code to partial rankings is given in the Supplementary. Concerning the Kemeny embedding, we recall that it encodes for any full ranking $\sigma \in \mathfrak{S}_K$, for each pair $i < j$, if $\sigma(i) < \sigma(j)$ i.e if $i \succ j$. It can thus be naturally extended to the case of incomplete and partial rankings. Indeed, for any partial ranking $\widetilde{\sigma}$, we propose to map it to the vector:

$$\phi(\widetilde{\sigma}) = (sign(\widetilde{\sigma}(i) - \widetilde{\sigma}(j)))_{1 \leq i < j \leq n}$$

where each coordinate can now take its value in $\{-1, 0, 1\}$ (instead of $\{-1, 1\}$ for full rankings). For any incomplete ranking $\bar{\sigma}$, we also propose to fill the missing entries (missing comparisons) in the embedding with zeros. This can be interpreted as setting the probability that $i \succ j$ to $1/2$ for a missing comparison between $(i, j)$. The Lehmer and Kemeny embedding can thus be naturally extended to partial or/and incomplete rankings since they encode *relative* information about the positions of the items. The Hamming embedding, in contrast, encodes the absolute positions of the items and is then much trickier to extend to partial or incomplete rankings.

## 5. Computational and theoretical analysis

### 5.1 Theoretical guarantees

In this section, we give some statistical guarantees when learning predictors by following the steps described section 3. To this end, we build on recent results in the framework of Least Squares Loss Surrogate by Ciliberto et al. (2016). Consider one of the embeddings on permutations presented in the previous section $\phi$, which defines a loss $\Delta$ as in (2). Let $c_\phi = \max_{\sigma \in \mathfrak{S}_K} \|\phi(\sigma)\|$. We will denote by $s^*$ a minimizer of the true risk (1), $g^*$ a minimizer of the surrogate risk (3), and $d$ a decoding function as (8)[5]. Given an estimator $\widehat{g}$ of $g^*$ from Step 1, i.e. a minimizer of the empirical surrogate risk (6) we can then consider in Step 2 an estimator $\widehat{s} = d \circ \widehat{g}$. The following theorem reveals how the performance of the estimator $\widehat{s}$ we propose can be related to a solution $s^*$ of (1) for the considered embeddings.

**Theorem 1** *The excess risks of the proposed predictors are linked to the excess surrogate risks as:*
*(i) For the loss (2) defined by the Kemeny and Hamming embedding $\phi_\tau$ and $\phi_H$ respectively:*

$$\mathcal{E}(d \circ \widehat{g}) - \mathcal{E}(s^*) \leq c_\phi \sqrt{\mathcal{R}(\widehat{g}) - \mathcal{R}(g^*)}$$

*with $c_{\phi_\tau} = \sqrt{\frac{K(K-1)}{2}}$ and $c_{\phi_H} = \sqrt{K}$.*
*(ii) For the loss (2) defined by the Lehmer embedding $\phi_L$:*

$$\mathcal{E}(d \circ \widehat{g}) - \mathcal{E}(s^*) \leq \sqrt{\frac{K(K-1)}{2}} \sqrt{\mathcal{R}(\widehat{g}) - \mathcal{R}(g^*)} + \mathcal{E}(d \circ g^*) - \mathcal{E}(s^*)$$

The full proof is given in the Supplementary. Assertion (i) is a direct application of Theorem 2 in Ciliberto et al. (2016). In particular, it comes from a preliminary consistency result which shows that $\mathcal{E}(d \circ g^*) = \mathcal{E}(s^*)$ for both embeddings. Concerning the Lehmer embedding, it is not possible to apply their consistency results immediately; however a large part of the arguments of their proof is used to bound the estimation error for the surrogate risk, and we remain with an approximation error $\mathcal{E}(d \circ g^*) - \mathcal{E}(s^*)$ resulting in Assertion (ii). In Remark 3 in the Supplementary, we give several insights about this approximation error in (ii). Firstly we show that it can be upper bounded by $4c_{\phi_L} \|d_L\| \mathcal{E}(s^*)$ (with $\|.\|$ the operator norm). Then, we explain how this term results from using $\phi_L$ in the learning procedure. The Lehmer embedding thus have weaker statistical guarantees, but has the advantage of being more computationnally efficient, as we explain in the next subsection.

Notice that for Step 1, one can choose a consistent regressor with vectorial values $\widehat{g}$, i.e such that $\mathcal{R}(\widehat{g}) \to \mathcal{R}(g^*)$ when the number of training points tends to infinity. Examples of such methods. that we use in our experiments to learn $\widehat{g}$, are the k-nearest neighbors (kNN) or kernel ridge regression (see Micchelli and Pontil (2005)) methods whose consistency have been proved (see Chapter 5 in Devroye et al. (2013) and Caponnetto and De Vito (2007)). In this case the control of the excess of the surrogate risk $\mathcal{R}(\widehat{g}) - \mathcal{R}(g^*)$ implies the control of $\mathcal{E}(\widehat{s}) - \mathcal{E}(s^*)$ where $\widehat{s} = d \circ \widehat{g}$ by Theorem 1.

---

5. Note that $d = \phi_L^{-1} \circ d_L$ for $\phi_L$ and is obtained as the composition of two steps for $\phi_\tau$ and $\phi_H$: solving an optimization problem and compute the inverse of the embedding.

| Embedding | Step 0 | Step 2 |
|---|---|---|
| $\phi_\tau$ | $\mathcal{O}(K^2N)$ | NP-hard |
| $\phi_H$ | $\mathcal{O}(KN)$ | $\mathcal{O}(K^3N)$ |
| $\phi_L$ | $\mathcal{O}(KN)$ | $\mathcal{O}(KN)$ |

| Regressor | Step 1 | Prediction |
|---|---|---|
| kNN | $\mathcal{O}(1)$ | $\mathcal{O}(Nm)$ |
| Ridge | $\mathcal{O}(N^3)$ | $\mathcal{O}(Nm)$ |

Table 1: Embeddings and regressors complexities.

### 5.2 Algorithmic complexity

We now discuss the algorithmic complexity of our approach. We recall that $K$ is the number of items/labels whereas $N$ is the number of samples in the dataset. For a given embedding $\phi$, the total complexity of our approach for learning decomposes as follows. Step 0: map the dataset of rankings with $\phi$, Step 1: compute the Least squares surrogate minimization (6). Then the prediction consists in mapping new inputs to a Hilbert space using the solution of Step 1 and then Step 2: solve the preimage problem (7). We report the algorithmic complexity of each Step and the one of the regressor prediction phase in Table (1). We denote by $m$ the dimension of such predictions. The complexity of a predictor corresponds to the worst complexity across all steps. The complexities resulting from the choice of an embedding and a regressor are summarized Table 1. The Lehmer embedding with kNN regressor thus provides the fastest theoretical complexity of $\mathcal{O}(KN)$ at the cost of weaker theoretical guarantees. The fastest methods previously proposed in the litterature typically involved a sorting procedure at prediction Cheng et al. (2010) leading to a $\mathcal{O}(NKlog(K))$ complexity. In the experimental section we compare our approach, with the former (denoted as Cheng PL), but also with the label wise decomposition approach in Cheng and Hüllermeier (2013) (Cheng LWD) involving a kNN regression followed by a projection on $\mathfrak{S}_K$ computed in $\mathcal{O}(K^3N)$, and the more recent Random Forest Label Ranking (Zhou RF) Zhou and Qiu (2016). In their analysis, if $d_\mathcal{X}$ is the size of input features and $D_{\max}$ the maximum depth of a tree, then RF have a complexity in $\mathcal{O}(D_{\max}d_\mathcal{X}K^2N^2)$.

### 6. Numerical Experiments

We now evaluate the performance of our approach on standard benchmarks. We present the results obtained with two regressors : Kernel Ridge regression (Ridge) and k-Nearest Neighbors (kNN). Both regressors were trained with the three embeddings presented in Section 4. We adopt the same setting as Cheng et al. (2010) and report the results of our predictors in terms of mean Kendall's $\tau$:

$$k_\tau = \frac{C-D}{n(n-1)/2} \quad \begin{cases} C : \text{number of concordant pairs between 2 rankings} \\ D : \text{number of discordant pairs between 2 rankings} \end{cases}$$

from five repetitions of a ten-fold cross-validation (c.v.). Note that $k_\tau$ is an affine transformation of the Kendall's tau distance $\Delta_\tau$ mapping on the $[-1, 1]$ interval. We also report the standard deviation of the resulting scores as in Cheng and Hüllermeier (2013). The parameters of our regressors were tuned in a five folds inner c.v. for each training set. We report our parameter grids in the Supplementary. The Kemeny and Lehmer embedding based approaches are competitive with the state of the art methods on these benchmarks datasets. The Hamming based methods give poor results in terms of $k_\tau$ but become the best choice when measuring the mean Hamming distance between predictions and ground truth (see Table 3 in the Supplementary). The Supplementary presents additional results which show that our method remains competitive with the state of the art.

Table 2: Mean Kendall's $\tau$ coefficient on benchmark datasets

|  | authorship | glass | iris | vehicle | vowel | wine |
|---|---|---|---|---|---|---|
| kNN Kemeny | **0.94**±0.02 | 0.85±0.06 | 0.95±0.05 | 0.85±0.03 | 0.85±0.02 | 0.94±0.05 |
| kNN Lehmer | 0.93±0.02 | 0.85±0.05 | 0.95±0.04 | 0.84±0.03 | 0.78±0.03 | 0.94±0.06 |
| ridge Hamming | -0.00±0.02 | 0.08±0.05 | -0.10±0.13 | -0.21±0.03 | 0.26±0.04 | -0.36±0.03 |
| ridge Lehmer | 0.92±0.02 | 0.83±0.05 | **0.97**±0.03 | 0.85±0.02 | 0.86±0.01 | 0.84±0.08 |
| ridge Kemeny | **0.94**±0.02 | 0.86±0.06 | **0.97**±0.05 | **0.89**±0.03 | **0.92**±0.01 | 0.94±0.05 |
| Cheng PL | **0.94**±0.02 | 0.84±0.07 | 0.96±0.04 | 0.86±0.03 | 0.85±0.02 | **0.95**±0.05 |
| Cheng LWD | 0.93±0.02 | 0.84±0.08 | 0.96±0.04 | 0.85±0.03 | 0.88±0.02 | 0.94±0.05 |
| Zhou RF | 0.91 | **0.89** | **0.97** | 0.86 | 0.87 | **0.95** |

## 7. Conclusion

This paper introduces a novel framework for label ranking, which is based on the theory of Surrogate Least Square problem for structured prediction. The embeddings we propose come along with theoretical guarantees and efficient algorithms, and prove their performance on real-world datasets. To go forward, extensions of our methodology to predict partial and incomplete rankings are to be investigated. In particular, the framework of prediction with abstention should be of interest.

### Acknowledgements

This research is supported by the chair Machine Learning for Big Data of Telecom ParisTech.

# 8. Supplementary material

## 8.1 Proof of Theorem 1

We borrow the notations of Ciliberto et al. (2016) and recall their main result Theorem 2. They firstly exhibit the following assumption for a given loss $\Delta$, see Assumption 1 therein:

**Assumption 1.** There exists a separable Hilbert space $\mathcal{F}$ with inner product $\langle .,.\rangle_\mathcal{F}$, a continuous embedding $\psi : \mathcal{Y} \to \mathcal{F}$ and a bounded linear operator $V : \mathcal{F} \to \mathcal{F}$, such that:

$$\Delta(y, y') = \langle \psi(y), V\psi(y') \rangle_\mathcal{F} \quad \forall y, y' \in \mathcal{Y} \qquad (20)$$

**Theorem 2** *Let $\Delta : \mathcal{Y} \to \mathcal{Y}$ satisfying Assumption 1 with $\mathcal{Y}$ a compact set. Then, for every measurable $g : \mathcal{X} \to \mathcal{F}$ and $d : \mathcal{F} \to \mathcal{Y}$ such that $\forall h \in \mathcal{F}$, $d(h) = \mathrm{argmin}_{y \in \mathcal{Y}} \langle \phi(y), h \rangle_\mathcal{F}$, the following holds:*

*(i) Fisher Consistency: $\mathcal{E}(d \circ g^*) = \mathcal{E}(s^*)$*

*(ii) Comparison Inequality: $\mathcal{E}(d \circ g) - \mathcal{E}(s^*) \leq 2c_\Delta \sqrt{\mathcal{R}(g) - \mathcal{R}(g^*)}$*

*with $c_\Delta = \|V\| \max_{y \in \mathcal{Y}} \|\phi(y)\|$.*

Notice that any discrete set $\mathcal{Y}$ is compact and $\phi : \mathcal{Y} \to \mathcal{F}$ is continuous. We now prove the two assertions of Theorem 1.

*Proof of Assertion(i) in Theorem 1.* Firstly, $\mathcal{Y} = \mathfrak{S}_K$ is finite. Then, for the Kemeny and Hamming embeddings, $\Delta$ satisfies Assumption 1 with $V = -id$ (where $id$ denotes the identity operator), and $\psi = \phi_K$ and $\psi = \phi_H$ respectively. Theorem 2 thus applies directly.

*Proof of Assertion(ii) in Theorem 1.* In the following proof, $\mathcal{Y}$ denotes $\mathfrak{S}_K$, $\phi$ denotes $\phi_L$ and $d = \phi_L^{-1} \circ d_L$ with $d_L$ as defined in (17). Our goal is to control the excess risk $\mathcal{E}(s) - \mathcal{E}(s^*)$.

$$\mathcal{E}(s) - \mathcal{E}(s^*) = \mathcal{E}(d \circ \widehat{g}) - \mathcal{E}(s^*)$$
$$= \underbrace{\mathcal{E}(d \circ \widehat{g}) - \mathcal{E}(d \circ g^*)}_{(A)} + \underbrace{\mathcal{E}(d \circ g^*) - \mathcal{E}(s^*)}_{(B)}$$

Consider the first term (A).

$$\mathcal{E}(d \circ \widehat{g}) - \mathcal{E}(d \circ g^*) = \int_{\mathcal{X} \times \mathcal{Y}} \Delta(d \circ \widehat{g}(x), \sigma) - \Delta(d \circ g^*(x), \sigma) dP(x, \sigma)$$
$$= \int_{\mathcal{X} \times \mathcal{Y}} \|\phi(d \circ \widehat{g}(x)) - \phi(\sigma)\|_\mathcal{F}^2 - \|\phi(d \circ g^*(x)) - \phi(\sigma)\|_\mathcal{F}^2 dP(x, \sigma)$$
$$= \underbrace{\int_\mathcal{X} \|\phi(d \circ \widehat{g}(x))\|_\mathcal{F}^2 - \|\phi(d \circ g^*(x))\|_\mathcal{F}^2 dP(x)}_{(A1)} +$$
$$\underbrace{2 \int_\mathcal{X} \langle \phi(d \circ g^*(x)) - \phi(d \circ \widehat{g}(x)), \int_\mathcal{Y} \phi(\sigma) dP(\sigma, x) \rangle dP(x)}_{(A2)}$$

The first term (A1) can be upper bounded as follows:

$$\int_\mathcal{X} \|\phi(d \circ \widehat{g}(x))\|_\mathcal{F}^2 - \|\phi(d \circ g^*(x))\|_\mathcal{F}^2 dP(x) \leq \int_\mathcal{X} \langle \phi(d \circ \widehat{g}(x)) - \phi(d \circ g^*(x)), \phi(d \circ \widehat{g}(x)) + \phi(d \circ g^*(x)) \rangle_\mathcal{F} dP(x)$$
$$\leq 2c_\Delta \int_\mathcal{X} \|\phi(d \circ \widehat{g}(x)) - \phi(d \circ g^*(x))\|_\mathcal{F} dP(x)$$
$$\leq 2c_\Delta \|\phi \circ d\| \int_\mathcal{X} \|g^*(x) - \widehat{g}(x)\|_\mathcal{F}^2 dP(x)$$

with $c_\Delta = \max_{\sigma \in \mathcal{Y}} \|\phi(\sigma)\|_\mathcal{F} = \frac{(n-1)(n-2)}{2}$ and $\|\phi \circ d\|$ the operator norm. For the second term (A2), we can actually follow the proof of Theorem 12 in Ciliberto et al. (2016) and we get:

$$\int_\mathcal{X} \langle \phi(d \circ g^*(x)) - \phi(d \circ \widehat{g}(x)), \int_\mathcal{Y} \phi(\sigma) dP(\sigma, x) \rangle dP(x) \leq 2c_\Delta \sqrt{\mathcal{R}(\widehat{g}) - \mathcal{R}(g^*)}$$

Consider the second term (2). By Lemma 8 in Ciliberto et al. (2016), we have that:

$$g^*(x) = \int_\mathcal{Y} \phi(\sigma) dP(\sigma|x) \tag{21}$$

and then:

$$\mathcal{E}(d \circ g^*) - \mathcal{E}(s^*) = \int_{\mathcal{X} \times \mathcal{Y}} \|\phi(d \circ g^*(x)) - \phi(\sigma)\|_\mathcal{F}^2 - \|\phi(s^*(x)) - \phi(\sigma)\|_\mathcal{F}^2 dP(x, \sigma)$$

$$\leq \int_{\mathcal{X} \times \mathcal{Y}} \langle \phi(d \circ \widehat{g}(x)) - \phi(s^*(x)), \phi(d \circ \widehat{g}(x)) + \phi(s^*(x)) - 2\phi(\sigma) \rangle_\mathcal{F} dP(x, \sigma)$$

$$\leq 4c_\Delta \int_\mathcal{X} \|\phi(d \circ g^*(x)) - \phi(s^*(x))\|_\mathcal{F} dP(x)$$

$$\leq 4c_\Delta \int_\mathcal{X} \|d_L \circ g^*(x)) - d_L \circ \phi(s^*(x))\|_\mathcal{F} dP(x)$$

where the last inequality comes from the fact that $d = \phi_L^{-1} \circ d_L$ and $\phi(s^*(x)) \in \mathcal{C}_K$ so $d_L \circ \phi(s^*(x)) = \phi(s^*(x))$. Then we can plug (21) in the right term:

$$\mathcal{E}(d \circ g^*) - \mathcal{E}(s^*) \leq 4c_\Delta \|d_L\| \int_\mathcal{X} \|\int_\mathcal{Y} \phi(\sigma) dP(\sigma|x) - \phi(s^*(x))\|_\mathcal{F} dP(x)$$

$$\leq 4c_\Delta \|d_L\| \int_{\mathcal{X} \times \mathcal{Y}} \|\phi(\sigma) - \phi(s^*(x))\|_\mathcal{F} dP(x)$$

$$\leq 4c_\Delta \|d_L\| \mathcal{E}(s^*)$$

**Remark 3** *As proved in Theorem 19 in Ciliberto et al. (2016), since the space of rankings $\mathcal{Y}$ is finite, $\Delta_L$ necessarily satisfies Assumption 1 with some continuous embedding $\psi$. If the approach we developed was relying on this $\psi$, we would have consistency for the minimizer $g^*$ of the Lehmer loss (16). However, the choice of $\phi_L$ is relevant because it yields a pre-image problem with low computational complexity.*

### 8.2 Lehmer embedding for partial rankings

An example, borrowed from Li et al. (2017) illustrating the extension of the Lehmer code for partial rankings is the following:

| | 1 | 2 | 3 | 4 | 5 | 6 | 7 | 8 | 9 |
|---|---|---|---|---|---|---|---|---|---|
| e | 1 | 2 | 3 | 4 | 5 | 6 | 7 | 8 | 9 |
| $\widetilde{\sigma}$ | 1 | 1 | 2 | 2 | 3 | 1 | 2 | 3 | 3 |
| $\sigma$ | 1 | 2 | 4 | 5 | 7 | 3 | 6 | 8 | 9 |
| $c_\sigma$ | 0 | 0 | 0 | 0 | 0 | 3 | 1 | 0 | 0 |
| $IN$ | 1 | 2 | 1 | 2 | 1 | 3 | 3 | 2 | 3 |
| $c_{\widetilde{\sigma}}$ | 0 | 0 | 0 | 0 | 0 | 3 | 1 | 0 | 0 |
| $c'_{\widetilde{\sigma}}$ | 0 | 1 | 0 | 1 | 0 | 5 | 3 | 1 | 2 |

where each row represents a step to encode a partial ranking.

### 8.3 Additional experimental results

**Details concerning the parameter grids.** We first recall our notations for vector valued kernel Ridge regression. Let $\mathcal{H}_K$ be a vector-valued Reproducing Kernel Hilbert Space associated to an operator-valued kernel $K : \mathcal{X} \times \mathcal{X} \to \mathcal{L}(\mathbb{R}^n)$ where $\mathcal{L}(\mathbb{R}^n)$ is the set of bounded linear operators from $\mathbb{R}^n$ to $\mathbb{R}^n$. A vector valued Ridge regressor $g$ is obtained by solving the optimization problem:

$$\min_{g \in \mathcal{H}_K} \sum_{k=1}^{N} \|g(x_k) - \phi(\sigma_k)\|^2 + \lambda \|h\|^2_{\mathcal{H}_K} \tag{22}$$

The solution of this problem is unique and admits an expansion: $\widehat{g}(.) = \sum_{i=1}^{N} K(x_i, .)c_i$ (see Micchelli and Pontil (2005)). Moreover, it has the following closed-form solution:

$$\widehat{g}(.) = \psi_x(.)(K_x + \lambda I_{nN})^{-1} Y_N \tag{23}$$

where $K_x$ is the $N \times N$ block-matrix, with each block of the form $K(x_k, x_l)$, $Y_N$ is the vector of all stacked vectors $\phi(\sigma_1), \ldots, \phi(\sigma_N)$, $\psi_x$ is the matrix composed of the block matrices $[K(., x_1), \ldots, K(., x_N)]$ and $I_{nN}$ is the $nN \times nN$ dimensional identity matrix. In all our experiments, we used a decomposable gaussian kernel $K(x, y) = \exp(-\gamma \|x - y\|^2) I_n$ where $I_n$ is the $n \times n$ identity matrix. The bandwith $\gamma$ and the regularization parameter $\lambda$ were chosen in the set $\{10^{-i}, 5 \cdot 10^{-i}\}$ for $i \in 0, \ldots, 5$ during the gridsearch cross-validation steps. For the k-Nearest Neighbors experiments, we used the euclidean distance and the neighborhood size was chosen in the set $\{1, 2, 3, 4, 5, 8, 10, 15, 20, 30, 50\}$.

**Experimental results.** We report additional results in terms of rescaled Hamming distance (if $K$ is the number of object to rank, the rescaled Hamming distance expression is $d_{H_K}(\sigma, \sigma') = \frac{d_H(\sigma, \sigma')}{K^2}$) on the datasets presented in the paper and in terms of Kendall's $\tau$ coefficient on other datasets. All the results have been obtained in the same experimental conditions: ten folds cross-validation are repeated five times with the parameters tuned in a five folds inner cross-validation. The results presented in Table 3 correspond to the mean normalized Hamming distance between the prediction and the ground truth. Whereas Hamming based embeddings led to very low results on the task measured using the Kendall's $\tau$ coefficient, they outperform other embeddings for the Hamming distance minimization problem. In the table (4), we show that Lehmer and Hamming based embeddings stay competitive

Table 3: rescaled Hamming distance (lower is better)

|  | authorship | glass | iris | vehicle | vowel | wine |
|---|---|---|---|---|---|---|
| kNN Kemeny | 0.05±0.01 | 0.07±0.02 | 0.04±0.03 | 0.08±0.01 | 0.07±0.01 | 0.04±0.03 |
| kNN Lehmer | 0.05±0.01 | 0.08±0.02 | 0.03±0.03 | 0.10±0.01 | 0.10±0.01 | 0.04±0.03 |
| kNN Hamming | 0.05±0.01 | 0.08±0.02 | 0.03±0.03 | 0.08±0.02 | 0.07±0.01 | 0.04±0.03 |
| ridge Kemeny | 0.06±0.01 | 0.08±0.03 | 0.04±0.03 | 0.08±0.01 | 0.08±0.01 | 0.04±0.03 |
| ridge Lehmer | 0.05±0.01 | 0.09±0.03 | **0.02**±0.02 | 0.10±0.01 | 0.08±0.01 | 0.09±0.04 |
| ridge Hamming | **0.04**±0.01 | **0.06**±0.02 | **0.02**±0.02 | **0.07**±0.01 | **0.05**±0.01 | **0.04**±0.02 |

on other standard benchmark datasets.

Table 4: Kendall's $\tau$ coefficient on additional datasets

|  | bodyfat | calhousing | cpu-small | pendigits | segment | wisconsin | fried |
|---|---|---|---|---|---|---|---|
| kNN Lehmer | **0.23**±0.01 | 0.22±0.01 | 0.40±0.01 | **0.94**±0.00 | 0.95±0.01 | **0.49**±0.00 | 0.85±0.02 |
| kNN Kemeny largedata | **0.23**±0.06 | 0.33±0.01 | **0.51**±0.00 | **0.94**±0.00 | 0.95±0.01 | **0.49**±0.04 | 0.89±0.00 |
| Cheng PL | **0.23** | 0.33 | 0.50 | **0.94** | 0.95 | 0.48 | 0.89 |
| Zhou RF | 0.185 | **0.37** | **0.51** | **0.94** | **0.96** | 0.48 | **0.93** |